\documentclass[10pt,journal,compsoc]{IEEEtran}
\ifCLASSOPTIONcompsoc
  \usepackage[nocompress]{cite}
\else
  \usepackage{cite}
\fi
\ifCLASSINFOpdf
\else
\fi

\usepackage{graphicx}
\usepackage{amsmath}
\usepackage{amssymb}
\usepackage{booktabs}
\usepackage{multirow}
\usepackage{bbm}
\usepackage{adjustbox}

\usepackage{amsmath,amsfonts,bm}

\def\eqref#1{equation~\ref{#1}}

\def\1{\bm{1}}

\DeclareMathAlphabet{\mathsfit}{\encodingdefault}{\sfdefault}{m}{sl}
\SetMathAlphabet{\mathsfit}{bold}{\encodingdefault}{\sfdefault}{bx}{n}

\usepackage[usenames,dvipsnames]{xcolor}
\usepackage{xcolor}
\usepackage{pifont}
\usepackage{algorithm}
\usepackage{listings}
\usepackage{comment}
\usepackage{makecell}
\usepackage{lipsum}

\renewcommand{\paragraph}[1]{\vspace{1.25mm}\noindent\textbf{#1}}

\usepackage[pagebackref,breaklinks,colorlinks]{hyperref}

\usepackage[capitalize]{cleveref}
\crefname{section}{Sec.}{Secs.}
\Crefname{section}{Section}{Sections}
\Crefname{table}{Table}{Tables}
\crefname{table}{Tab.}{Tabs.}

\begin{document}
%

\title{A Survey on Masked Autoencoder for Self-supervised Learning in Vision and Beyond}

\author{Chaoning~Zhang, Chenshuang~Zhang, Junha~Song, John Seon Keun~Yi, Kang~Zhang, In So~Kweon%
\IEEEcompsocitemizethanks{
\IEEEcompsocthanksitem The authors are with KAIST.\protect\\
E-mail: chaoningzhang1990@gmail.com
}
}

%
%

\markboth{Journal of \LaTeX\ Class Files,~Vol.~14, No.~8, August~2015}
{Zhang \MakeLowercase{\textit{et al.}}: A Survey on Masked Autoencoder for Self-supervised Learning}
%



\IEEEtitleabstractindextext{%
\begin{abstract}
Masked autoencoders are scalable vision learners, as the title of MAE~\cite{he2022masked}, which suggests that self-supervised learning (SSL) in vision might undertake a similar trajectory as in NLP. Specifically, generative pretext tasks with the masked prediction (e.g., BERT) have become a de facto standard SSL practice in NLP. By contrast, early attempts at generative methods in vision have been buried by their discriminative counterparts (like contrastive learning); however, the success of mask image modeling has revived the masking autoencoder (often termed denoising autoencoder in the past). As a milestone to bridge the gap with BERT in NLP, masked autoencoder has attracted unprecedented attention for SSL in vision and beyond. This work conducts a comprehensive survey of masked autoencoders to shed insight on a promising direction of SSL. As the first to review SSL with masked autoencoders, this work focuses on its application in vision by discussing its historical developments, recent progress, and implications for diverse applications.
\end{abstract}

}

\maketitle

\IEEEdisplaynontitleabstractindextext

%
\IEEEpeerreviewmaketitle

\IEEEraisesectionheading{\section{Introduction}\label{sec:introduction}}

%
%
%
%

\IEEEPARstart{D}{eep} learning~\cite{lecun2015deep} has revolutionized artificial intelligence in the past decade. Early developments focused on the architecture design with scalable size like increasing model depth, from AlexNet~\cite{krizhevsky2012imagenet} to VGG~\cite{simonyan2014very} and ResNet~\cite{he2016deep}. In recent years, the attention has gradually shifted from designing better models to solving the data-hungry issue in deep learning. For example, ImageNet~\cite{deng2009imagenet} with more than one million labeled images has become a typical benchmark dataset for vision models, and vision transformer (ViT)~\cite{dosovitskiy2021an} is reported to demand hundreds of times more labeled images. A common way to perform satisfactorily with a relatively small labeled dataset is to pre-train the model on another larger dataset, which is widely known as transfer learning. Self-supervised learning (SSL)~\cite{he2020momentum,chen2020simple}, outperforming its supervised counterpart for visual pre-training, has attracted significant attention. 

With the advent of contrastive SSL in 2018, joint-embedding methods have become a dominant visual pre-trainign framework; however, this status has been recently challenged by the success of a generative method termed masked image modeling (MIM)~\cite{bao2022beit}. BEiT~\cite{bao2022beit} adopts a mask-then-predict strategy to train the model with the target visual tokens generated by an off-the-shelf tokenizer. The tokenizer is pretrained by a discrete variational autoencoder (dVAE)~\cite{ramesh2021zero}, and therefore BEiT can be seen as a two-stage training of denoising autoencoder~\cite{vincent2008extracting}. Furthermore, an end-to-end masked autoencoder in the vision is proposed in MAE~\cite{he2022masked}, which has attracted unprecedented attention.

As the term suggests, a masked autoencoder is an autoencoder with masked prediction, \textit{i.e.} predicting a property of masked input from unmasked input content. It is worth mentioning that masked autoencoder is not something new in unsupervised visual pretraining. Dating back to 2008, an early work~\cite{vincent2008extracting} predicted masked pixels from unmasked ones but was referred to as denoising autoencoder~\cite{vincent2008extracting,vincent2010stacked}. A similar investigation was conducted again in 2016 with the task of image inpainting~\cite{pathak2016context}. Its reviving success in recent MAE~\cite{he2022masked}, outperforming joint-embedding methods, inspires numerous works to understand its success in vision and to apply it in various applications, such as video, point cloud, and graph.

The reason for high popularity of masked autoencoder in visual pretraining is that a similar generative SSL framework termed masked language modeling (like BERT~\cite{devlin2019bert}) has been widely used in NLP. In other words, the success of masked autoencoder in vision paves a path that SSL in vision``\textit{may now be embarking on a similar trajectory as in NLP}"~\cite{he2022masked} by generative pretext task with masked prediction. Moreover, since NLP and computer vision are two dominant branches in modern AI, many researchers believe that masked autoencoder might be the future for SSL.

To this end, this work conducts a comprehensive survey of masked autoencoders in SSL. This survey covers its application with various data types; however, it focuses on understanding its reviving success in vision. Note that autoencoder-based masked prediction started to become a de facto standard practice in language understanding in 2018/2019~\cite{devlin2019bert}; thus, it is less relevant to discuss it in the 2020s. Moreover, it is the success of masked autoencoder in vision that shows visual SSL can embark on the same path as that in language, which somewhat revolutionizes visual SSL and then inspires the investigation of masked autoencoder in a wide range of applications. With masked autoencoder in vision as the focus, this survey mainly contains three parts. (1) Sec.~\ref{sec:from_NLP} summarizes its historical development and relation with masked language modeling; (2) Sec.~\ref{sec:MIM_understanding} discusses the masked modeling principle in vision and the understanding of its success from various perspectives. (3) Sec.~\ref{sec:to_others} summarizes its implications on pre-training in diverse applications beyond natural images. To facilitate discussion without ambiguity, we include a terminology section (\textit{i.e.} Sec.~\ref{sec:terminology}) to discuss essential terms in this survey.

\textbf{Message to the readers.} This survey will be updated on a regular basis to reflect the dynamic progress of masked autoencoder in its development. Since masked autoencoder is a fast-evolving field, and we might not be able to grasp all recent development. Therefore, we encourage researchers to contact us to inform us with their new works, either published ones or arXiv ones, on this topic. Those new works will be included and discussed in the revised version.

\section{Background and terminology} \label{sec:terminology}

\textbf{Generative SSL \textit{v.s.} discriminative SSL.} In self-supervised learning, modelling methods can be roughly categorized into: discriminative or generative. Generative SSL typically relies on an autoeocnder that consists of encoding (\textit{i.e.} mapping an input to a latent representation with an encoder) and decoding (\textit{i.e.} generating the input from the latent representation with an decoder)~\cite{ng2011sparse}. Discriminative SSL typically follows its supervised couterpart to design a discriminative loss. Without ground-truth labels, a discrimiantive pretext task can be designed as solving jigsaw puzzles~\cite{noroozi2016unsupervised} or predicting rotation~\cite{gidaris2018unsupervised}. Later, the trend of discriminative visual SSL shifts from such geometry-based prediction to joint-embedding emthods~\cite{zhang2022how,zhang2022dual,jing2021understanding}.

\textbf{Denoising autoencoder \textit{v.s.} masked autoencoder.} As a classical generative SSL method, denoising autoencoder is a class of autoencoders that reconstruct the original clean input from a corrupted input~\cite{vincent2008extracting,vincent2010stacked}. Note that \textit{denoising} in this context (and in this whole survey) refers to reconstruction from general corruption (including but not limited to noise). Since \textit{masked prediction} refers to the practice of predicting a property of masked input from unmasked input, it can be seen as a form of denoising process~\cite{yi2022masked}. This predicted property can be the original input~\cite{yi2022masked}, handcrafted feature~\cite{wei2022masked}, or latent representation~\cite{baevski2022data2vec}. Since masked prediction is a form of denoising process and thus masked autoencoder can be seen as a form of general denoising autoencoder. In this work, we use MAE exclusively to refer to the method in~\cite{he2022masked} \textit{not} as shorthand for masked autoencoder to avoid confusion.

\textbf{Masked autoencoding \textit{v.s.} masked modeling.} Masked prediction can be used to both generative and discriminative modeling methods. However, the term masked \textit{X} modeling, namely masked modeling on X-type data, often refers to the generative case, such as masked \textit{language} modeling~\cite{devlin2019bert}, masked \textit{image} modeling~\cite{xie2022simmim}, masked \textit{point} modeling~\cite{yu2022point}. Motivated by its success in generative modeling, a few works~\cite{baevski2022data2vec,assran2022masked,yi2022masked} have also applied masked prediction in discriminative SSL frameworks, demonstrating competitive performance. In other words, masked modeling is not necessarily masked autoencoding. 
Take image data, for example, MSN~\cite{assran2022masked} and data2vec~\cite{baevski2022data2vec} can be categorized as masked image modeling but not masked autoencoding since their model architectures are decoder-free. In this work, we still perceive BEiT~\cite{bao2022beit} as a variant of masked autoencoder even though it decouples the pretext task of masked prediction from autoencoder training.

\begin{figure*}[t]\centering
\includegraphics[width=1\linewidth]{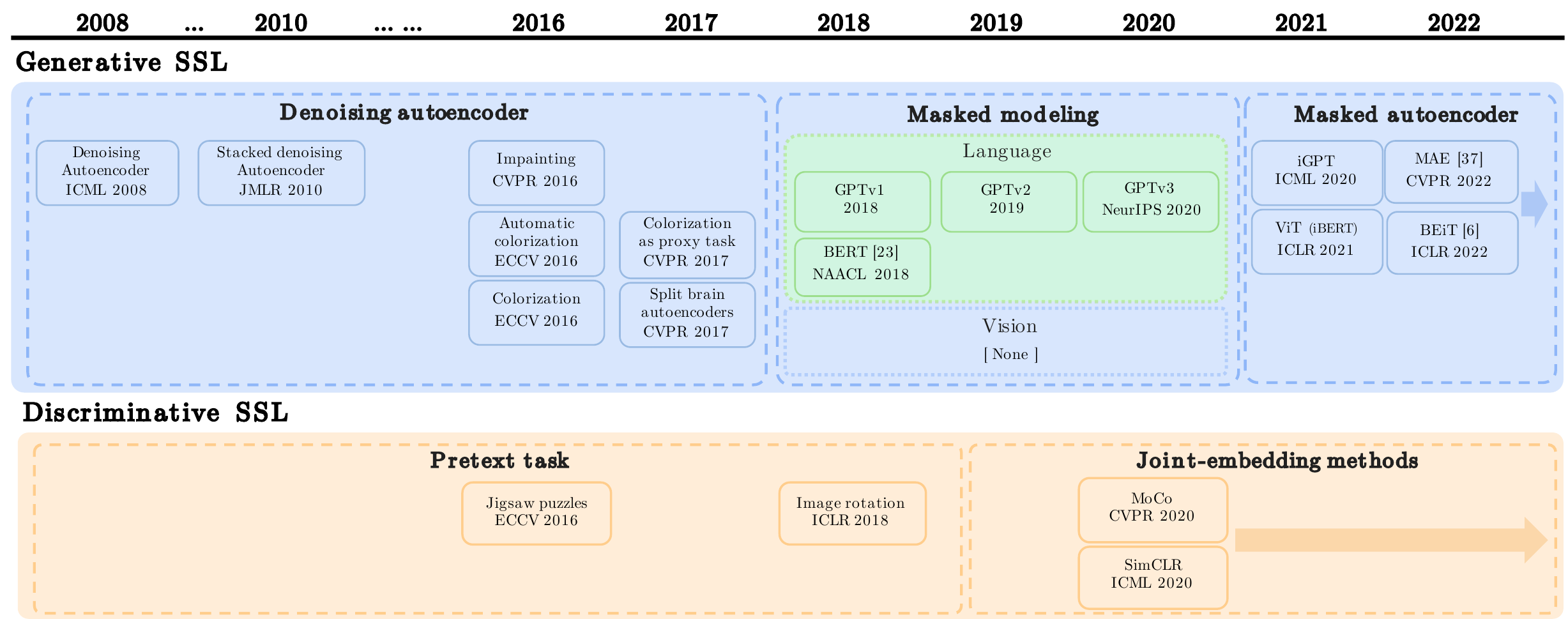}\\
\caption{Timeline of Visual SSL}. 
\label{fig:timeline}
\vspace{-1em}
\end{figure*}

\section{Masked autoencoding: NLP to vision}
\label{sec:from_NLP}

NLP and (computer) vision are two dominant research fields for artificial intelligence. Despite the difference in the data types and downstream tasks, vision and language communities have often inspired each other. Towards a unified understanding of language and image, it is interesting to ask whether they can adopt a similar backbone architecture and training strategy. For the backbone architecture, the advent of vision Transformer (ViT) in~\cite{dosovitskiy2021an} and its application to various vision tasks demonstrate that Transformers~\cite{vaswani2017attention} can serve as a unified backbone architecture for both language and vision. Numerous works have further attempted to bridge the gap in their SSL training strategies. 

\subsection{NLP and vision followed different SSL paths} \label{sec:two_paths}
\textbf{Generative SSL in NLP.} In NLP there exist two leading language models: GPT~\cite{radford2018improving,radford2019language,brown2020language} and BERT~\cite{devlin2019bert}. They are both based on the transformer architecture but with notable differences~\cite{bilogur2020notes}: GPT works by predicting the next word based on previous words and thus is autoregressive in nature, while BERT uses the entire surrounding context of words all at once. In essence, they both remove a portion
of the data and predict the removed content, and they can be both perceived to rely on masked prediction as the pretext task.

\textbf{Discriminative SSL in vision.}
Before 2018, there was an active investigation of unsupervised visual pretraining with both generative and discriminative modeling. Jigsaw and rotation prediction is designed as the pretext task on the discriminative side, while inpainting and colorization have been actively investigated on the generative side (see Sec.\ref{sec:xxx}). Since 2018, joint-embedding methods, namely aligning the embedded representations of augmented views of the same image~\cite{jing2021understanding}, have demonstrated substantial performance boost over prior generative methods. Contrastive learning~\cite{wu2018unsupervised,oord2018representation}, which makes the representations of positive samples close and those of negative samples far from each other, has emerged as a dominant visual SSL method, especially after the advent of MoCo~\cite{he2020momentum} and SimCLR~\cite{chen2020simple}. Negative-free (\textit{i.e.} non-contrastive) joint-embedding methods have also been investigated~\cite{zbontar2021barlow,chen2021exploring,zhang2022how}, demonstrating comparable performance of contrastive learning methods. A unified perspective on contrastive learning and negative-free joint-embedding is extensively discussed in~\cite{zhang2022how,zhang2022dual}.

\subsection{Is generative SSL suitable for vision?} \label{sec:xxx}

\textbf{Very early attempts.} Images have spatial and channel dimensions, and therefore we can either predict masked spatial patches from unmasked ones~\cite{vincent2008extracting,vincent2010stacked,pathak2016context} or predict masked channels from unmasked ones~\cite{larsson2016learning,larsson2017colorization,zhang2016colorful,zhang2017split}. A standard autoencoder takes an image as the input and reconstructs it after the information passes through a low-dimensional bottleneck layer. Without corrupting the input, the encoder focuses on content compression instead of extracting semantically meaningful representations. Denoising autoencoder was proposed in ~\cite{vincent2008extracting,vincent2010stacked} to perform masked autoencoding in the spatial dimension by randomly masking some pixels. To make it a harder task to avoid learning only low-level representation,~\cite{pathak2016context} proposed feature learning by inpainting, \textit{i.e.} to fill in large missing areas of the image and thus prevent hints from nearby
pixels. Later,~\cite{larsson2016learning,zhang2016colorful} showed that masked channel prediction yielded superior performance on downstream tasks, especially for dense semantic segmentation, since it keeps the spatial content. They were further improved in~\cite{larsson2017colorization,zhang2017split}. The investigation in this direction has been less active since the emergence of contrastive learning (see discussion in Sec.\ref{sec:two_paths}).

\textbf{Inspiration from NLP.} The above investigation~\cite{vincent2008extracting,vincent2010stacked,pathak2016context,larsson2016learning,larsson2017colorization,zhang2016colorful,zhang2017split} was before 2017. With GPT and BERT emerging in 2018/2019 to show the success of masked prediction in language understanding, a natural question is: can we transfer the success of masked modeling from language to image? iGPT~\cite{chen2020generative} is the first successful attempt in this direction; however, as highlighted in~\cite{chen2020igpt_blog}, their work is for proof-of-concept and cannot be used in practice due to two reasons: (1) it takes two orders higher pre-training compute than contrastive methods and (2) it performs worse than contrastive methods based on CNN. As the first attempt to replace CNN with a transformer in vision,~\cite{dosovitskiy2021an} identified that the success of transformer in NLP tasks stems from excellent scalability and self-supervised pre-training. Since the self-supervised pre-training practice in~\cite{dosovitskiy2021an} mimicked the masked language modeling task in BERT, we call it iBERT in analogy to iGPT entending GPT from language to vision. iBERT performs a masked patch prediction for visual SSL. However, this preliminary investigation of ViT for SSL also shows inferior performance over joint-embedding methods. This challenge was finally broken by BEiT~\cite{bao2022beit} as well as MAE~\cite{he2022masked} (see Sec.\ref{sec:MIM_understanding} for their details).

\subsection{Summary and remark}
\textbf{Summary.} Figure~\ref{fig:timeline} shows the overall timeline for the development of unsupervised visual pretraining (including GPT and BERT for NLP). Interestingly, unsupervised visual pretraining started with generative SSL in 2008. Its reviving attempt in 2016 and 2017 was then buried by discriminative SSL, especially after the advent of joint-embedding methods. However, with the inspiration from NLP, generative SSL with masked prediction comes back again.

\begin{table}[htb] \centering
\vspace{-10pt}
\caption{Comparison of denosing autoencoder~\cite{vincent2008extracting} and masked autoencoder~\cite{he2022masked}}
\label{tab:denoising_masked}
\resizebox{0.5\textwidth}{!}{ 
\begin{tabular}{cccc} 
\hline
& denoising autoencoder~\cite{vincent2008extracting} & masked autoencoder~\cite{he2022masked}  \\ 
\hline
Training dataset        &      MNIST       &     ImageNet  \\ \hline
Model Architecture      &      CNN       &     ViT  \\ \hline
Corruption size       &      pixels       &     patches  \\ \hline
Corruption ratio       &      maximum 50\%       &     patches  \\ 
\hline
\end{tabular}}
\end{table}
\textbf{Remark.} Early denoising autoencoder~\cite{vincent2008extracting} and recent masked autoencoder~\cite{he2022masked} both attempt to reconstruct a clean input from a corrupted one, precisely predicting masked input content from unmasked input content. Despite high similarity regarding pretext task, the masked autoencoder introduced in~\cite{he2022masked} differs from early denoising autoencoder~\cite{vincent2008extracting} in numerous ways, which are summarized in Table~\ref{tab:denoising_masked}.

\section{Masked autoencoder for image modeling}
\label{sec:MIM_understanding}
As discussed in Sec.\ref{sec:from_NLP}, iGPT and iBERT have shown the possibility of transferring the pretext task of masked prediction from language to image data. However, their performance is inferior to joint-embedding methods and thus has caught less attention. BEiT is the first to show the success of autoencoder-based masked prediction outperforming DINO, a SOTA joint-embedding method. Therefore, this section starts with introducing BEiT with its improved variants.

\subsection{BEiT and its improved variants} \label{sec:beit}

\begin{figure}[!htbp]\centering
\includegraphics[width=0.98\linewidth]{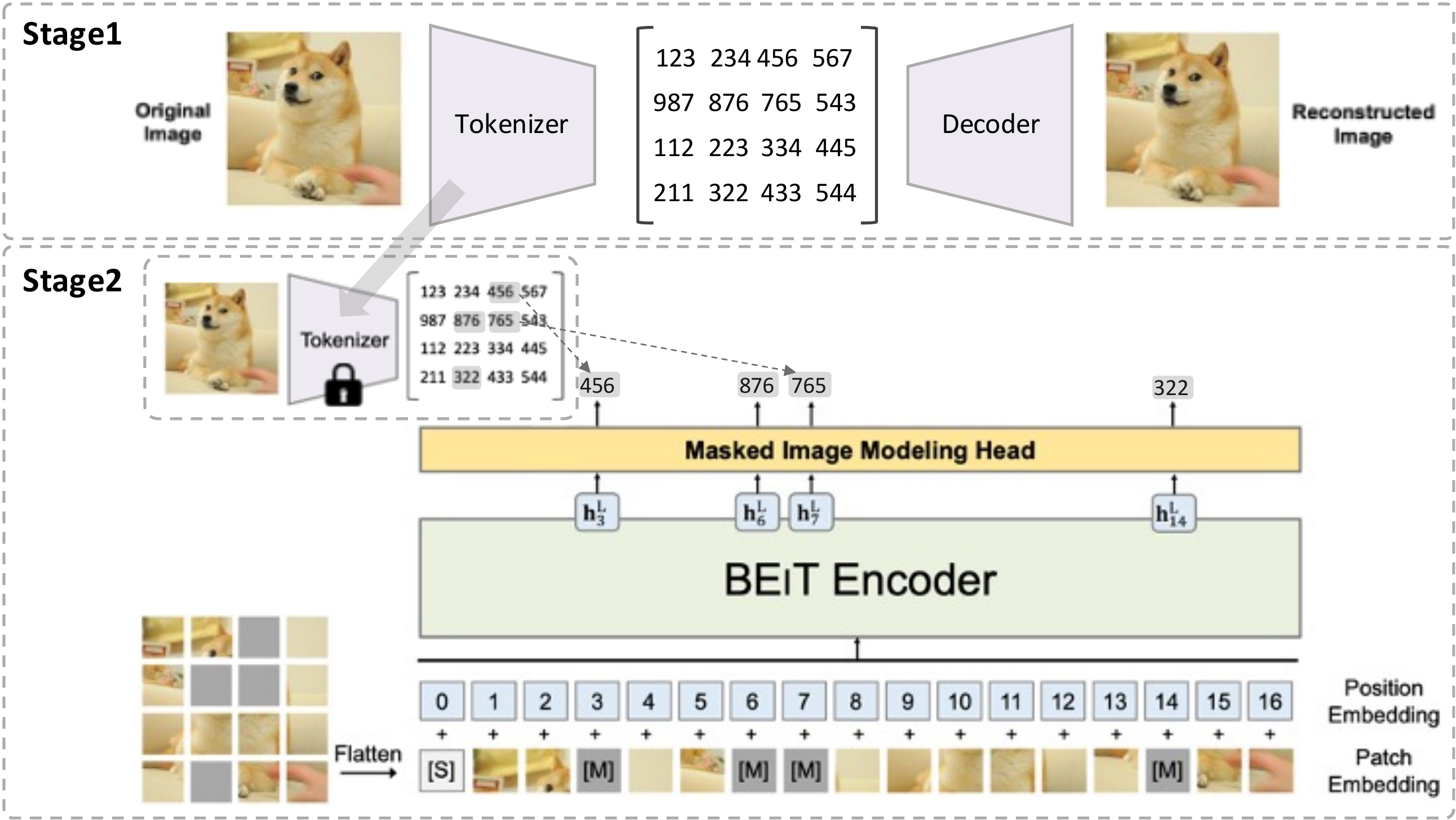}\\
\caption{Overview of BEIT pre-training. The figure is edited from~\cite{bao2022beit}.
}
\label{fig:beit}
\end{figure}

\textbf{BEiT.} The overview of BEiT is shown in Figure~\ref{fig:beit}. In contrast to iBERT~\cite{dosovitskiy2021an} that directly reconstructs the masked patches, BEiT mimicks BERT~\cite{devlin2019bert} to reconstruct visual tokens. Since Image patches do not have off-the-shelf tokens as words in the language, BEiT trains an image tokenizer via discrete variational autoencoder (dVAE)~\cite{ramesh2021zero} before the second-step masked image modeling where the tokenizer is used to guide the learning of BEiT encoder (note that decoder is unused). Specifically, the tokenizer takes the original image, and the BEiT encoder takes a corrupted image, including unmasked patches and masked patches. Then, it outputs the visual tokens of masked patches to match the corresponding visual tokens from the tokenizer (staying fixed in this process). BEiT is the first to show that masked image modeling has downstream task performance superior to SOTA contrastive DINO~\cite{caron2021emerging}. Despite its success, it remains unknown whether directly predicting masked image patches as in iBERT~\cite{dosovitskiy2021an} might be a simpler alternative.   

BEiT~\cite{bao2022beit} consists of two stages: token-based MIM as the main stage and tokenizer training as the preparation stage. Multiple works~\cite{dong2021peco,li2022mc,chen2022context} have followed this two-stage approach by either improving the tokenizer-based MIM process or seeking an alternative tokenizer.

\textbf{Tokenizer-based MIM.} mc-BEiT~\cite{li2022mc} attempts to effectively utilize the visual tokenizer generated by dVAE. Specifically, it observes that unlike linguistic vocabulary consisting of discrete words, the image tokenizer is continuous. Under visual discretization, visual patches with similar semantics can have different token IDs, and visual patches with different semantics can have the token ID, which is not desired. Therefore, mc-BEiT recasts the MIM in BEiT from a single-choice classification problem to a multiple-choice one by softening the training objective from a hard-label cross-entropy loss to a soft-label one. 
Following BEiT~\cite{bao2022beit}, CAE~\cite{chen2022context} first trains a image tokenizer via dVAE to generate target visual tokens. BEiT performs the encoding and decoding role implicitly and simultaneously, while CAE performs the two tasks explicitly and separately. A key component realizes this termed \textit{latent contextual regressor} to introduce alignment between the representations of masked patches and unmasked ones. The CAE encoder \textit{exclusively} focuses on feature extraction without making predictions for masked patches. The CAE encoder exploits the full representation capability by letting the latent contextual regressor handle the prediction pretext task.

\textbf{Better target tokenizer.} PeCo~\cite{dong2021peco} identifies that the visual tokenizer generated by dVAE~\cite{ramesh2021zero} does not consider semantic level. PeCo adds the distance between deep visual features as an extra loss to enforce perceptual similarity between the original image and the reconstructed image to make the target visual tokens more semantically meaningful. For studying masked prediction,~\cite{wei2022masked} follows the two-stage approach as BEiT and investigates various target tokenizers. Interestingly, it is found that handcrafted HOG features~\cite{dalal2005histograms} achieve a competitive performance, suggesting a target tokenizer generated by dVAE might be unnecessary. However, HOG is only compatible with visual data and limits its applications in other data modalities.

\subsection{End-to-end masked autoencoder}
A drawback of the two-stage methods is that their approach relies on a pretrained dVAE to generate originally continuous but \textit{intentionally discretized} target visual tokens~\cite{yi2022masked}, and thus is not end-to-end. In essence, BEiT separates masked prediction from autoendoer training, which leaves room for improving effectiveness and efficiency. To this end, MAE~\cite{he2022masked} experiments with end-to-end training of masked autoencoder. We highlight that SimMIM~\cite{xie2022simmim} has conducted a very similar investigation. MAE and SimMIM appear on arXiv concurrently (MAE being one week earlier) and are both accepted at CVPR'2022. Here, we summarize the two works and compare their nuanced difference. 

\begin{figure}[!htbp]\centering
\includegraphics[width=.9\linewidth]{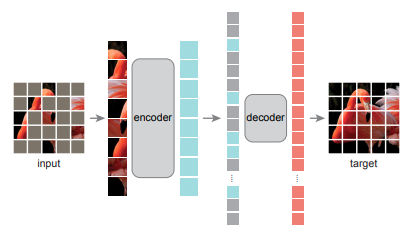}\\
\caption{Overview of a masked autoencoder with the figure borrowed from the original work MAE~\cite{he2022masked}.
}
\label{fig:mae}
\vspace{-1em}
\end{figure}

\textbf{MAE.} The overview of MAE~\cite{he2022masked} is shown in Figure~\ref{fig:mae}. MAE revisits the pretext task of predicting masked patches. Specifically, their proposed MAE~\cite{he2022masked} directly predicts masked patches from the unmasked ones with a simple loss of mean squared error (MSE). Moreover, the masking ratio is set to 75\%, which is significantly higher than that in BERT (typically 15\%)~\cite{devlin2019bert} or prior MIM (20\% to 50\%)~\cite{chen2020generative, dosovitskiy2021an,bao2022beit}. The ablation findings support such a high masking ratio is beneficial for fine-tuning and linear probing. It is worth mentioning that this also motivates a recent work to experiment with a higher masking rate in masked language modeling for higher effectiveness~\cite{wettig2022should}. To save computation, the encoder of MAE only operates on the unmasked patches. Moreover, MAE designs an asymmetric encoder-decoder architecture with a lightweight decoder. With the above technical tricks, their proposed simple MAE is (3 $\times$ or more) faster than BEiT~\cite{bao2022beit} while achieving superior performance. 

\begin{figure}[!htbp]\centering
\includegraphics[width=.99\linewidth]{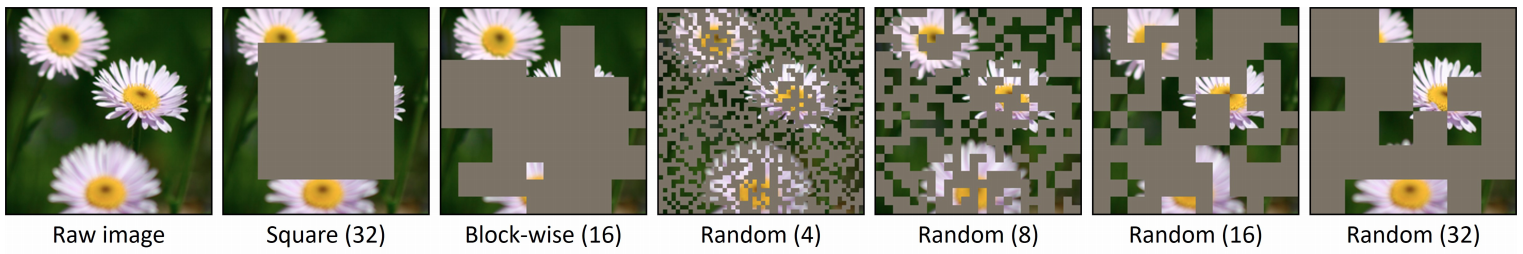}\\
\caption{Various masking strategies in SimMIM with the figure borrowed from the original paper~\cite{xie2022simmim}.
}
\label{fig:simmim}
\vspace{-1em}
\end{figure}

\textbf{SimMIM.} Independently and concurrently, a similar architecture termed Simple Masked Image Modeling (SimMIM) is proposed in ~\cite{xie2022simmim}, where similar findings are reported. Specifically, SimMIM confirms that directly predicting the pixels as in MAE performs no worse than other methods with complex design, such as tokenization, clustering, or discretization. It is also found that moderately increasing the patch size (32, for instance) is beneficial for a more powerful pretext task. A high masking ratio is also confirmed in MAE to be helpful for performance, especially for a relatively small patch size. Moreover, as shown in Figure~\ref{fig:simmim}, SimSIM investigates multiple masking strategies, such as square, block-wise, and random. Their best performance is achieved with the random masking strategy, which is the same as that in MAE.

\textbf{Difference between MAE and SimMIM.} One of their non-trivial differences lies in the position of masked patch tokens. Specifically, masked patch tokens are adopted as the input of decoder and decoder in MAE~\cite{he2022masked} and SimMIM~\cite{xie2022simmim}, respectively. With the pretext task of masked prediction, the autoencoder in MAE and SimMIM fulfills two roles: representation encoding (for unmasked patches) and pretext prediction (for masked patches). With both masked and unmasked patches as the input, the encoder of  SimMIM~\cite{xie2022simmim} simultaneously performs representation encoding and pretext prediction, due to which the decoder can be designed as simple as a single layer. By contrast, the encoder in MAE~\cite{he2022masked} exclusively realizes representation encoding, leaving the role of pretext prediction to the decoder. As a result, MAE still relies on a transformer decoder, as reported in~\cite{he2022masked}, even though it does not need to be as heavy as the encoder. Due to this, MAE achieves significantly higher linear probing accuracy than SimMIM; however, this superiority diminishes with finetuning. For example, with ViT-B as the backbone on ImageNet, SimMIM achieves a finetuning performance of 83.8\%, slightly higher than the reported 83.6\% for MAE. Another merit of MAE by feeding only the unmasked patches into the encoder is its higher efficiency, especially when the masking ratio is high. Unlike SimMIM with Swin-B as the default backbone, MAE is not compatible with hierarchical ViT (like Swin~\cite{wang2021pvt,liu2021swin}). The reason for its incompatibility and solutions to address them are discussed in the following.

\subsection{Towards improving efficiency}
Despite the impressive performance, a significant bottleneck of masked autoencoder for visual SSL is that it requires large computation. In this section, we introduce multiple works that attempt to improve the efficiency of masked autoencoders from roughly two perspectives: (1) hierarchical structure and (2) input manipulation.

\textbf{Hierarchical structure.} 
Since ViT~\cite{dosovitskiy2021an} used in MAE has a crucial issue that decreasing the patch size will quadratically increase computing resources, hierarchical ViT (hViT)~\cite{wang2021pvt,liu2021swin} was introduced. Specifically, Swin and PVT~\cite{liu2021swin, wang2021pvt} use a shrinking pyramid structure with additional tricks such as shifted windows~\cite{liu2021swin} to learn local feature correlations or spacial reduction attention~\cite{wang2021pvt} to reduce computation in the attention layer used to further improve performance. Unfortunately, it is not intuitive to adapt hViT to enable MAE pre-training since the local window attention used in hViT is challenging to handle randomly masked patches as in MAE.

Several works~\cite{huang2022green, li2022uniform, zhang2022hivit} attempt to boost the power of hViTs while achieving efficiency in MAE. Huang et al. ~\cite{huang2022green} present a unique masking strategy called group window attention that gathers unmasked patches into several equal-sized groups to perform masked attention. Their method, based on a Swin transformer, combines the multi-scale feature learnability of hViT and the efficiency of masked image modeling by making the hierarchical transformer compatible with MAE. 
Similarly, Uniform Masking MAE (UM-MAE)~\cite{li2022uniform} introduced a two-stage sampling and masking process. The proposed Uniform Masking strategy first uniformly samples a quarter (25\%) of patches in each block, then further masks random patches on top of the sampled patches. The first step maintains similar elements across the non-overlapped local windows, while the second step makes the self-supervisory task more challenging by avoiding shortcuts for pixel reconstruction from neighboring low-level features. HiViT~\cite{zhang2022hivit} proposes a new hViT architecture to substitute window attention layers in Swin~\cite{liu2021swin} with MLP layers which enables masking as in MAE. The above works~\cite{huang2022green, li2022uniform, zhang2022hivit} achieve comparable performance to the baselines (MAE, SimMIM) while requiring less training time as well as less GPU memory.

\textbf{Input manipulation.}
Several methods attempt to improve the efficiency of MAE by changing the input. Specifically, they aim to reduce the input size by attending to small windows~\cite{chen2022efficient} or objects in the image~\cite{wu2022object}. These methods reduce the required computation while achieving comparable or better downstream task performance. 

Local masked reconstruction (LoMaR)~\cite{chen2022efficient} is inspired from the fact that local information is enough for reconstructing masked patches. Instead of relying on the entire image for mask reconstruction, a number of small windows with 7x7 patches are sampled to restrict attention to local regions. LoMaR achieves higher downstream task performance faster compared with MAE. It excels on high-resolution images since the required compute increase linearly with image size where it is quadratic for MAE. ObjMAE~\cite{wu2022object} achieves input efficiency by dropping non-object patches and learning object-wise representations. A class activation map (CAM)~\cite{zhou2016learning} is used to identify a rough object region, and the object regions are masked and used as input for the MAE. ObjMAE reduces the pre-training compute cost by 72\% while achieving comparable performance to MAE, which masks the whole scene. MixMIM~\cite{liu2022mixmim} takes a slightly different approach: to replace an image's masked tokens with tokens from another image. The mixed image is then fed into a encoder then the decoder reconstructs the two original images. Because of the absence of uninformative masked tokens, \cite{liu2022mixmim} is not only able to be suitable for hierarchical ViTs such as Swin~\cite{liu2021swin} but also achieves stronger results efficiently compared to existing MIM works.

\subsection{Various perspectives on the success of masked autoencoder in vision} 
To explain why BEIT~\cite{bao2022beit} helps the finetuning on downstream tasks, its authors analyze the self-attention map and show that BEiT distinguishes semantic regions using self-attention heads without any task-specific supervision. Moreover,~\cite{he2022masked} shows that an MAE, pretrained with a masking ratio of 75\%, infers complex and holistic reconstructions even when 95\% of pixels are masked, suggesting it learns various concepts, \textit{i.e.}, \ semantics. The authors of MAE~\cite{he2022masked} ``hypothesize that this behavior occurs through a rich hidden representation inside the MAE". Given that the masked and reconstructed visual patches are not semantic entities as words in languages, this behavior is somewhat unexpected and is hypothesized to occur ``by way of a rich hidden representation"~\cite{he2022masked}. However, which component in masked autoencoder makes the model learn such a ``rich hidden representation" remains unclear. Numerous works have investigated from various perspectives for a better understanding of its success.

\textbf{Backbone perspective: Is masked autoencoder compatible with CNN?} With ViT~\cite{dosovitskiy2021an} as the default backbone in MAE~\cite{he2022masked}, a natural question is whether masked autoencoder works only with a transformer backbone instead of CNN. Since CNN cannot tackle the masked inputs and positional embedding directly, multiple works~\cite{gao2022convmae,fang2022unleashing,li2022architecture,fang2022corrupted} have attempted to unify ViT and CNN in a compatible masked autoencoder framework. Inspired by the observation that early convolutions help transformers see better~\cite{xiao2021early}, ConvMAE~\cite{gao2022convmae} utilizes hybrid convolution-transformer architectures: convolution blocks at early stages and transformer blocks at later stages are in charge of high-resolution token embedding and low-resolution token embedding, respectively. 
Towards a unified framework of MIM with both transformer and CNN architecture,~\cite{fang2022corrupted} proposes corrupted image modeling (CIM), which replaces the input images artificially masked in MIM with a corrupted image generated by a trainable generator (BEiT). Therefore, the reconstruction task in MIM can be extended to either generative or discriminative objectives trained by a ViT or CNN enhancer. CIM is the first to unify ViT and CNN in a non-Siamese framework and yields compelling results in vision benchmarks. More recently, it has been highlighted in~\cite{li2022architecture} that the success of masked image modeling can be agnostic to the architecture. The proposed Architecture Agnostic Masked Image Modeling framework (A$^2$MIM) is compatible with ViT and CNN in a unified way~\cite{li2022architecture}. It is found in~\cite{li2022architecture} that the success of masked autoencoder lies in learning middle-level patch interaction, which is agnostic to architecture choices. Early attempts of CNN-based inpainting~\cite{pathak2016context} resembles masked autoencoder but focuses on reconstruction task with low-level interactions, which causes higher feature uncertainty~\cite{li2022architecture}.

\textbf{Data perspective: does masked autoencoder require a very large dataset?} A popular belief regarding the benefit of transfer learning comes from pretraining on a much larger dataset than the target dataset. Challenging this belief,~\cite{el2021large} investigates whether self-supervised pretraining on a smaller dataset can yield the same benefit. The fact that their investigation is performed with ViT-based masked autoencoder makes it more interesting because, compared with its CNN, ViT is found to require much more samples~\cite{dosovitskiy2021an}. Interestingly,~\cite{dosovitskiy2021an} shows that pretraining masked autoencoder (either BEiT or SplitMakk~\cite{el2021large} ) on 1\% of ImageNet dataset achieves comparable transfer
performance to the iNaturalist-2019 dataset as pretraining on full ImageNet dataset. By contrast, prior DINO~\cite{caron2021emerging} is much more sensitive to the data size (as well as the data type). More recently,~\cite{xie2022data} performed a comprehensive study on data scaling (from 10\% of ImageNet to full ImageNet-22K) on masked autoencoder models of various sizes ranging from 49 million to 1 billion parameters. It shows that MIM is also demanding on larger data, especially for larger models with longer training epochs~\cite{xie2022data}. Beyond size, some works have also investigated domain issues in data and found that they can be alleviated by training the masked autoencoder on images of mixed-style~\cite{yang2022domain} or multi-tasks~\cite{bachmann2022multimae}.

\textbf{Denoising perspective: Does masked autoencoder benefit from other corruptions?} Given that masked autoencoder is a class of denoising autoencoder, \cite{tian2022beyond} investigates a general question: are there other effective image degradation methods beyond masking for effective visual pretraining? Five methods, namely zoom-in, zoom-out, distortion, blurring, and de-colorizing, have been investigated, and they are found to perform better than None (\textit{i.e.}, \ no pretraining), suggesting a unified denoising perspective on the success of masked autoencoder. Nonetheless, blurring and de-colorizing perform worse than other degradation methods with spatial transformation because they cause image style shift from the pretext task to the downstream task. Among them, zoom-in performs the best and is complementary with masking to further boost the performance. In contrast to existing spatial masking,~\cite{xie2022masked} also investigates frequency masking by predicting masked high-frequency from the unmasked low-frequency content, or vice versa, demonstrating competitive performance. Moreover, super-resolution, deblur, and denoise have also been investigated but they yield inferior performance.

\textbf{Theoretical perspective: can masked autoencoder be explained with rigorous mathematics?}
Towards a mathematical understanding,~\cite{cao2022understand} was the first to propose a unified theoretical
framework for understanding masked autoencoder in vision. Particularly, each image’s embedding in MAE can be interpreted not as a 2D pixel grid but as a learned basis function in certain Hilbert spaces. Moreover, under a non-overlapping domain decomposition setting, the patch-based attention in ViT can be understood from the operator theoretic perspective of an integral kernel. With attention as the focus,~\cite{cao2022understand} further proves that the stability of internal representations and that masked latent representations are interpolated globally with an inter-patch topology. To understand why MAE helps in downstream tasks, based on an autoencoder of a two/one-layered CNN,~\cite{pan2022towards} theoretically shows that it can capture discriminative semantics in the pretraining dataset. Deviating from the focus on attention~\cite{cao2022understand}, it provides insight on what features MAE learns
and why MAE beats conventional supervised learning (SL). Particularly, MAE encoder captures all discriminative semantics in the pretraining dataset, including samples that have either single or multiple independent discriminative semantics, and therefore provably outperforms SL on downstream tasks.

\subsection{Relationship with joint-embedding methods} 

Before the success of masked autoencoder, visual self-supervised pretraining had been dominated by joint-embedding methods, either contrastive ones (~\cite{chen2020simple,chen2021mocov3}) or negative-free ones~\cite{grill2020bootstrap,caron2021emerging}. Thus, it is highly relevant to compare masked autoencoder with joint-embedding for visual self-supervised pretraining.

\subsubsection{Masked autoencoder and joint embedding: boosting each other}

An intriguing observation regarding their difference is as follows: compared with joint-embedding methods~\cite{chen2021mocov3,caron2021emerging}, masked autoencoders~\cite{he2022masked,xie2022simmim} have stronger finetuning performance on the downstream tasks but weaker linear probing accuracy. A popular understanding is that masked autoencoder lacks in learning semantically-meaningful features because it focuses on low-level patch match with a local loss~\cite{he2022masked,xie2022simmim}. On the other hand, high-level semantic features have the property of being robust to spatial transformation (like random crop) and style change (like color jittering)~\cite{misra2020self}, and thus joint embedding approaches adopt a global loss on the features after global average pooling to encourage the learned representation to be augmentation-invariant.

\textbf{Improving masked autoencoder with global loss.} SplitMask~\cite{el2021large} consists of three steps: split, inpaint, and match. The patches are divided into two disjoint subsets in the split step: $\mathcal{A}$ and $\mathcal {} $. For inpainting, it adopts a similar architecture as MAE in that a lightweight (shallow) ViT decoder is used to recover the masked patches from the representation of unmasked patches~\cite{el2021large}. What differentiates SplitMask~\cite{el2021large} from MAE~\cite{he2022masked} lies in the third match step, which encourages the global prediction of $\mathcal{A}$ and $\mathcal{B}$ subsets of patches to match each other. This global match aligns with the augmentation-invariant goal in joint-embedding approaches, thus making the representation more semantically meaningful.~\cite{tao2022siamese} improves MAE by combining it with joint-embedding approaches. Specifically, it predicts the masked tokens to match those from another augmented view to encourage semantic learning with an optional global loss.

\textbf{Improving joint-embedding methods with local loss.}

Multiple works in the above analysis show that the global loss in joint-embedding methods can be utilized to improve the semantic meaning of the learned representations. Intuitively, it is possible to improve the joint-embedding techniques by adding a local loss. For example, MST~\cite{li2021mst} extends the DINO framework by combining it with a masked prediction task. It is worth mentioning that MST~\cite{li2021mst} came out earlier than BEiT and MAE. More recently, RePre~\cite{wang2022repre} improves MoCO v3~\cite{chen2021mocov3} with a reconstruction loss by using a decoder to reconstruct the original image from the multi-hierarchy features in the encoder. ~\cite{wei2022contrastive} shows that their inferior fine-tuning performance can be significantly improved by a simple post-processing with feature distillation (FD). After FD, their representations are more suitable for optimization and thus finetuning friendly.

\subsubsection{Masked autoencoder and joint embedding: bridging their gap}

Masked autoencoder and joint-embedding perform masked prediction (predicting a property of masked patches from unmasked patches) and augmented alignment (aligning the embedded representation of different augmentations), respectively. From the perspective of the architecture component, the encoder training in masked autoencoder relies on a decoder, while that in joint-embedding uses a Siamese encoder for generating the self-supervision. Motivated by their success, multiple works have attempted masked prediction without a decoder, decoder-free MIM, which bridges the gap between joint-embedding and masked autoencoder for visual pretraining. 

\textbf{Decoder-free MIM.} Beyond masked autoencoder, decoder-free MIM can be seen as another line of simplifying BEiT from two stages to single stage. To keep the patch-level visual context, ConMIM~\cite{yi2022masked} follows the principle of designing the training objective to be masked patch prediction as in~\cite{bao2022beit}. Specifically, resembling MoCo~\cite{he2020momentum,chen2021mocov3}, ConMIM adopts a Siamese encoder, which is updated by the (student) encoder with EMA, as a teacher model to guide the training of the encoder. ConMIM~\cite{yi2022masked} feeds an unmasked image and a masked image of the same view into teacher and student encoders, respectively. The teacher encoder can be seen as a dynamic tokenizer as a static one in BEiT~\cite{bao2022beit}. Therefore, the embedded representations of masked patches are predicted to match the dynamic tokenizer corresponding to the same position~\cite{yi2022masked}. A similar teacher-student framework is adopted in MSN~\cite{assran2022masked} and data2vec~\cite{baevski2022data2vec}. In contrast to ConMIM~\cite{yi2022masked}, MSN~\cite{assran2022masked} adopts a global loss to encourage learning semantic-aware representation. CNN-based MSN has also been investigated in~\cite{jing2022masked}. It has also been demonstrated in data2vec~\cite{baevski2022data2vec} that this simple framework works well in the vision field and can be generalized to other data modalities, including speech and language. MSN~\cite{assran2022masked} works well for linear probing and few-shot learning but might be inferior to masked autoencoder for the finetuning performance on downstream tasks since patch-level visual context is discarded. To get the merits on both sides, iBOT~\cite{zhou2022ibot} adopts two losses: a local loss to distill in-view patch tokens and another global loss to distill between cross-view [CLS] tokens, which makes the target patch tokens more semantically-meaningful~\cite{zhou2022ibot}. More recently, AttMask~\cite{kakogeorgiou2022hide} shows that iBOT can be further improved by performing an attention-guided masking instead of random masking on the student side. Particularly, the teacher model indicates the attention with full image as the input and masking on the attended areas improves the performance on a variety of downstream tasks.

\textbf{Collapse issue.} A shared issue in the above decoder-free MIM methods~\cite{yi2022masked,assran2022masked,baevski2022data2vec} is the potential feature collapse, \textit{i.e.} outputting a constant for all inputs. They adopt different approaches to avoid this issue. For example, ConMIM~\cite{yi2022masked} adopts a contrastive loss with those feature representations corresponding to different positions in the same image as negative samples. MSN~\cite{assran2022masked} follows~\cite{caron2020unsupervised} to do cluster assignments, while data2vec~\cite{baevski2022data2vec} achieves this goal by carefully fine-tuning the hyperparameters like momentum coefficient in EMA and learning rate. Note that autoencoder-based MIM methods do not have the collapse issue by default. 

\section{Other Applications: Vision and Beyond} \label{sec:to_others}

Inspired by the success of MAE~\cite{he2022masked}, numerous works have applied masked autoencoder to various applications. We categorize them into two classes. The first class is related to vision, for which pure natural images have been extensively covered in the above section. Thus, this section covers its other aspects, including images with medical applications, images with temporal information, images with language. Going beyond vision, the second class focuses on different types of data, such as point clouds, graph, audio, reinforcement learning, etc.

\subsection{Vision related applications}

\subsubsection{Medical images}

Medical images are a class of data for medical analysis with data distribution different from natural images. Multiple works have shown that masked autoencoders can also work well in medical applications by either applying MAE directly to medical data~\cite{zhou2022self,chen2022masked} or improving the loss design~\cite{ly2022student,quan2022global,an2022masked,luo2022self}. With MAE~\cite{he2022masked} and SimMIM~\cite{xie2022simmim} as the architecture, ~\cite{zhou2022self} and ~\cite{chen2022masked} apply masked autoencoders directly in medical images, showing the effectiveness of masked autoencoder in medical applications, \textit{e.g.} CT images. Specifically, ~\cite{zhou2022self} shows that MAE pretrained on medical dataset achieves superior performance to its counterpart pretrained on ImageNet, which can be explained from the perspective domain shift. ~\cite{chen2022masked} shows that a moderately large patch size (32) achieves satisfactory performance, which aligns with the finding in~\cite{xie2022simmim}. There are also attempts to enhance MAE~\cite{he2022masked} by improving the loss~\cite{ly2022student,quan2022global,an2022masked,luo2022self}, including global loss and self-distillation loss. It is shown in~\cite{ly2022student,quan2022global} that an additional global loss on top of a local loss makes the representations more semantically meaningful for medical images, which resembles the principle in iBOT and SplitMask. ~\cite{an2022masked} views the output of MAE encoder as a bag of instances and aggregates the most informative tokens into global representation (slide-level) for further classification. To make full use of visible patches, Self-distillation MAE(SD-MAE) ~\cite{luo2022self} improves MAE by adding a self-distillation loss of visible patches between latent representations after encoding and decoding, which achieves competitive performance compared with contrastive methods. 

\subsubsection{Video}
Numerous works have applied SSL frameworks built on images to videos since videos are essentially a clip of sequential images. This trend is also observed after the success of masked autoencoders, with works in ~\cite{wang2022bevt,tan2021vimpac,wei2022masked} and  ~\cite{tong2022videomae,girdhar2022omnimae} applying videos to BEiT ~\cite{bao2022beit} and MAE~\cite{he2022masked} respectively. 

To learn spatial and temporal priors of videos in a decoupled way, BEVT ~\cite{wang2022bevt} proposes a two-stage solution that learns spatial representations with masked image modeling, then learns temporal representations with jointly masked 
image modeling and masked video modeling. BEVT achieves comparable or superior results to baseline methods on three video datasets. VIMPAC~\cite{tan2021vimpac} proposes a different single-stage masked video modeling method, which includes a block-wise masking strategy for videos and augmentation-free contrastive learning loss to learn the global features. Experimental results verify the effectiveness and scalability of the proposed VIMPAC. Both BEVT~\cite{wang2022bevt} and VIMPAC ~\cite{tan2021vimpac} rely on an external tokenizer which can be limited in compute-intensive video understanding scenarios. Therefore, ~\cite{wei2022masked} proposes to replace the tokens with features and investigates five types of features, among which hand-crafted HOG is found to work effectively and efficiently. 

Since MAE is found to be more simple yet effective than BEiT, the works in ~\cite{tong2022videomae,girdhar2022omnimae} follow the architecture of MAE for simplicity and efficiency. With a similar model architecture to MAE, VideoMAE~\cite{tong2022videomae} finds that 
it learns useful spatio-temporal structures with a very high masking ratio (90\% to 95\%) in tube masking strategy. Experimental results show that  VideoMAE~\cite{tong2022videomae} achieves impressive performance on tiny datasets. Similar investigation has also been investigated in~\cite{feichtenhofer2022masked} but with spacetime-agnostic random masking. Beyond video understanding for existing frames,~\cite{gupta2022maskvit} investigates masked visual modeling for future frame prediction. The gap between masked prediction for partial existing frames and full future frames is addressed by a variable masking ratio. OmniMAE~\cite{girdhar2022omnimae} extends MAE to a unified pre-training of image and video modalities. Trained on images and video with a single ViT encoder, OmniMAE achieves competitive performance on both image and video recognition benchmarks, outperforming models explicitly trained for a single modality.

\subsubsection{Vision and language}

Prior to masked autoencoder, contrastive learning is a popular approach to learn language and vision representations jointly. Contrastive Language-Image Pre-training (CLIP)~\cite{radford2021learning} is a pioneering work that propose learning images with language as supervision. By jointly learning an image and text encoder, CLIP takes the pair of image and text as a prediction target during contrastive pre-training and often achieves competitive results compared to fully supervised baselines. Other works extend of CLIP by adding self-supervision~\cite{mu2021slip}, data scaling~\cite{jia2021scaling} or enabling flexibility to the encoders~\cite{bao2022vlmo}.
Contrastive learning introduces sampling bias due to data augmentations and cannot tackle unpaired samples~\cite{geng2022multimodal}. To solve these problems,~\cite{geng2022multimodal} proposes Multimodal Masked Autoencoder (M3AE), which encodes a flexible mixture of inputs, including image-text pairs and image-only inputs. Experimental results show that  M3AE learns generalizable vision representations and unified information from images and languages. 
To investigate how to design an effective vision-language model with an end-to-end manner, Multimodal End-to-end TransformER (METER)~\cite{dou2022empirical} implements comprehensive experiments and analyses on multiple designs, including encoders, multimodal fusion module, pre-training objectives. However, adding MIM loss does not improve downstream task performance in their settings ~\cite{dou2022empirical}. ~\cite{bitton2021data} also investigates the masking strategies of text data in language-vision tasks, which improves performance on downstream tasks.

Moreover, ~\cite{lu2022unified} presents a unified task-agnostic model that can perform various vision and language tasks. It is able to tackle different tasks with a unified model without employing task-specific branches by tokenizing the inputs and outputs of every given task. A standard transformer encoder/decoder is pre-trained with masked language modeling and masked image modeling, then further trained on a large multi-task dataset that encompasses different language/vision tasks. 
Similarly, \cite{bao2022vl} proposes VL-BEiT that can tackle both monomodal and multimodal vision-language tasks. A single bidirectional multimodal transformer~\cite{bao2022vlmo} is pre-trained on mask prediction of monomodal (language, vision) and multimodal (image-text pair) data to be jointly optimized to different types of data. VL-BEiT achieves strong results on various vision-language benchmarks and image tasks. 

\subsection{Beyond vision}

\subsubsection{Point clouds}
Motivated by the success of BEiT in vision,~\cite{yu2022point} extends masked modeling strategy to point cloud with masked point modeling termed Point-BERT. Following BEiT, Point-BERT first trains a discrete VAE to generate discrete point tokens containing meaningful local information and then predicts the tokens of masked point patches from the unmasked point patches. With a pure transformer architecture surpassing carefully designed point cloud models, Point-BERT achieves 93.8\% accuracy on ModelNet40 and 83.1\% accuracy on the complicated setting of ScanObjectNN, suggesting the BERT-style pre-training technique also works for point cloud. Point-BERT relies on dVAE, which is trained by augmentation-based contrastive learning and thus is sophisticated. Moreover, the masked tokens from their inputs are processed as the input of Transformers, causing high compute and early leakage of location information~\cite{pang2022masked}. MaskPoint~\cite{liu2022masked} alleviates this issue by pre-training with a decoder to contrast masked points and noise. Moreover, Point-MAE~\cite{pang2022masked} follows MAE~\cite{he2022masked} to adopt a more straightforward approach to directly predict the locations of masked points. Resembling Point-MAE,~\cite{zhang2022point} proposes Point-M2AE, a Multi-scale MAE for point clouds. Different from Point-MAE, Point-M2AE adopts an encoder-decoder with pyramid architectures to capture both fine-grained and high-level semantics in a progressive manner. Accordingly, Point-M2AE also adopts a multi-scale masking strategy to yield visible patches consistent across scales. Moreover, a local spatial self-attention mechanism is also adopted to make the encoder focus on neighboring patterns. ~\cite{min2022voxel} has proposed Voxel-MAE to pre-train on large-scale point clouds to improve downstream 3D object detection. The key idea lies in dividing the point
clouds into voxel representations and classify whether they contain point clouds.

\subsubsection{Graph}
MGAE~\cite{tan2022mgae} is the first to investigate masked autoencoder for graphs. Prior to its advent, works on learning graph node representations in an unsupervised manner can be categorized into two classes: graph autoencoder and graph self-supervised learning (GSSL). They focus on designing effective encoder networks and advanced pretext tasks, respectively. Even though edge dropping and edge reconstruction are commonly adopted in both lines of investigations, masked autoencoding by recovering the masked edges from randomly masked input graph structure has never been explored until the advent of MGAE~\cite{tan2022mgae}. Following MAE~\cite{he2022masked}, MGAE operates only on convolution-based partial network structure (without masked edges). Moreover, the decoder is designed to capture the cross-correlation between an anchor edge's head and tail nodes. MGAE performs better or on par with graph autoencoder and GSSL. GMAE~\cite{chen2022graph} further investigates masked autoencoders for a graph with transformer instead of convolution. Another item that distinguishes them is that MGAE reconstructs masked edges, and GMAE reconstructs the features of masked nodes.~\cite{hou2022graphmae} also argues that rebuilding the features is more beneficial. Beyond empirical results with experimental trial-and-error, MaskGAE~\cite{li2022maskgae} further provides theoretical
justifications for the potential benefits of masked graph modeling.

\subsubsection{Reinforcement learning}
The auxiliary tasks and RL updates~\cite{haarnoja2018soft} are jointly trained in ~\cite{tao2022evaluating}, where the performances of ViT models is compared to that of a CNN-based RL method~\cite{laskin2020reinforcement}. The results indicate that CNN-based RAD~\cite{laskin2020reinforcement} performs better on most image-based deep RL tasks, but reconstruction-based ViT models~\cite{baevski2022data2vec,he2022masked} outperform RAD on some tasks. With ViT architecture, Xiao et. al~\cite{xiao2022masked} adopted pre-trained visual representations to train various motor control tasks. First, MAE~\cite{he2022masked} is used to learn visual representations from real-world images. Then, the encoder is freezed and the feature vector is used alongside propropceptive robot information to train task-specific motor cotrolling policies with model-free reinforcement learning\cite{schulman2017proximal}. The authors demonstrate that a single encoder can be used to learn various tasks without task-specific fine-tuning, and achieves superior performance compared to supervised baselines. Seo et. al~\cite{seo2022masked} demonstrate a similar approach, but show that convolutional \textit{feature} masking is more effective than pixel patch masking since it learns fine-grained features within patches.

\subsubsection{Audio}

Recent works learning audio representations create different input views by temporal relationships or data augmentations, which cannot provide information from the intact input. Inspired by the success of MAE in vision, ~\cite{niizumi2022masked,baade2022mae} apply masked autoencoders successfully on the masked audio spectrogram to learn audio representations from both time and frequency axes. Moreover,  ~\cite{baade2022mae} is trained jointly on discriminative and generative loss for instance-wise classification and reconstruction, respectively. 

\subsubsection{More diverse applications}
Masked autoencoder has also been attempted in more diverse applications. For example, ~\cite{zha2022time} experiments with masked autoencoder with extrapolator (ExtraMAE) to recover complex original time series signals from masked observations. Recognizing contrastive tabular-SSL does not sufficiently capture the underlying manifold due to the ad-hoc fashion of its augmentation design,~\cite{majmundar2022met} proposes Masked Encoding for Tabular data (MET). With the MAE~\cite{he2022masked} in vision as the baseline, MET adopts individual representation for each coordinate with an additional adversarial loss by considering the property of tabular data. Some works~\cite{fang2022corrupted,xu2022masked} have also adopted a pretrained masked autoencoder as an augmentation generator. For example, the reconstructed views from MAE are found to outperform hand-crafted augmentations (like scale, flip, and color jitter) in both supervised and semi-supervised setups~\cite{xu2022masked}.

\section{Conclusion}
This survey is the first to review the progress of masked autoencoder for SSL. We summarize the early attempts of masked autoencoder in vision and its relation with masked language modeling. With the focus on the reviving success of masked autoencoder in unsupervised visual pretraining, we summarize and compare the seminal methods as well as those follow-up works to improve them. We also provide insight on the success of masked autoencoder in vision from various perspectives, including backbone perspective, data perspective, denosing perspective and theoretical perspective. Finally, we summarize its application in vision and beyond. Preliminary summarization of the works in table format is provided in the appendix.

\bibliographystyle{IEEEtran}
\bibliography{bib_mixed}

\ifCLASSOPTIONcaptionsoff
  \newpage
\fi

\section{Appendix}
\begin{table*}[htb]\centering
\vspace{-10pt}
\caption{Summary of works with maksed autoencoder in vision.}
\label{tab:summary_of_models_image}
\resizebox{0.9\textwidth}{!}{ 
\begin{tabular}{cccccccccccccccc}
\hline 
model  &masking strategy &input type & prediction target & pretrain dataset(resolution)  & loss  & encoder & decoder     & Finetune Accuracy & arxiv time & publish status \\
\hline 
BEiT~\cite{bao2022beit} & block-wise & all patches & tokens &  ImageNet-1K(224)  & dVAE &ViT-B  & dVAE & 86.2\% &2021.06.15 &ICLR 2022\\
&&&&&& ViT-L & & 85.2\%\\
\hline 
MAE~\cite{he2022masked}& random & visible & pixel &  ImageNet-1K(224)  & MSE  & ViT-B  &  &  83.6\%  & 2021.11.11 & CVPR 2022\\
&&&&&& ViT-L  &8 block,width 512  &  85.9\%\\
&&&&&& ViT-H   && 86.9\%\\
\hline 
CAE ~\cite{chen2022context} &  block-wise &visible & tokens  & ImageNet-1K(224) & CE(MIM), MSE(align) &  VIT-S & 4 blocks    &82.0\% & 2022.02.07 & arxiv\\
&  &&&&&  VIT-B & 4 blocks & 83.9\% \\
&  &&&&&   VIT-L & 4 blocks   &86.3\% \\
\hline 
SimMIM~\cite{xie2022simmim} & random  &all patches   &  pixel &ImageNet-1K(224)  & &ViT-B & linear layer  &83.8\% &2021.11.18 & CVPR 2022\\
\hline 
Peco~\cite{dong2021peco} & random & all & token from proposed tokenizer & ImageNet-1K (800epoch) & same as BEiT & ViT-B & decoder & 84.5\% & 21.11.24 &  \\
\hline 
\textbf{With joint embedding} &  &  &  &  & \textbf{loss} & &  &  &  &  \\
\hline 
iBOT~\cite{zhou2022ibot}  & block-wise  & visible& tokens &  ImageNet-1K(224) & &VIT-B & None  & 87.8\% &2021.11.15 & ICLR2022\\
\hline 
SplitMask~\cite{el2021large} & block-wise & all patches & tokens   & ImageNet-1K(224)  & CE(MIM);InfoNCE & VIT-S & 2 blocks &81.5\% & 2021.12.20 & arxiv \\
  &  && & ImageNet-1K(224) &&   VIT-B & 2 blocks  & 83.6\% \\
\hline 
AttMask~\cite{kakogeorgiou2022hide} & attention-guided mask  & masked(student)\&unmasked(teacher) & teacher's feature & ImageNet-1K & distillation(KLdiv) & iBOT, ViT-S for attention & None & Incomparable & 22.05.23 & arXiv \\
 & =object-bias mask &  &  &  &  & &  &  &  &  \\
\hline 
ConMIM~\cite{yi2022masked} & random & masked(student)\&unmasked(teacher) & teacher's feature & ImageNet-1K(300) & InfoNCE & ViT-S & decoder & 82.0\% & 22.05.19 & arXiv \\
&  & & & & & VIT-B & &83.7\% \\
\hline 
SIM~\cite{tao2022siamese} & random & masked(student)\&unmasked(teacher) & teacher's feature & dataset & cosine similarity & ViT-B & decoder & Incomparable & 22.06.02 & arXiv \\
\hline 
\textbf{Contrastive leaning} & \textbf{with recon loss} &  &  &  & \textbf{loss} & &  &  &  &  \\
\hline 
MST~\cite{li2021mst} & no masking & all & pixel & ImageNet-1K (300) & DINO+l1 recon & DeiT-S & CNN & 76.9\% & 21.01.10 & NeurIPS 2021 \\
\hline 
Repre~\cite{wang2022repre} & random & all & pixel & ImageNet-1K (800) & DINO+l1 recon & ViT-S & CNN & 77.9\% & 22.01.18 & arXiv \\
& & & & ImageNet-1K (800) & & ViT-B & & 79.2\% & & \\ 

\hline 
\textbf{Efficient and Effective}  & \textbf{=fast pretraining} &  &  & (epoch) &  & &  &  &  &  \\
\hline 
LoMaR~\cite{chen2022efficient} & random windo\&patchs & all in windows & pixel & ImageNet-1K (300) & MSE & ViT-B & MLP (ReconHead) & 83.3\% & 22.06.01 & arxiv \\ 
& & & & ImageNet-1K (1600) & & & & 84.1\% & & \\ 
\hline 
ObjMAE~\cite{wu2022object} & random,object-aware & visible in object & pixel in object & ImageNet100 & MSE & sameMAE,ViT-B & sameMAE,ViT-B & Incomparable & 22.05.28 & arxiv \\
\hline 
Beyond Masking~\cite{tian2022beyond} & random mask+$\alpha$(ex,zoom-in..) & visible & pixel & ImageNet-1K (300) & MSE & sameMAE,ViT-B & sameMAE,ViT-B & 83.2\%(mask+zoomin) & 22.05.27 & arxiv \\
\hline 
\textbf{Lightweight architecture} \\ 
\hline 
GreenHierTransf~\cite{huang2022green} & group window attention & visible & Pixel & ImageNet-1K (800) & MSE & Swin-B & ViT & 83.7\% & 22.05.26 & arxiv \\
& & & & & & Swin-L & & 85.1\% & & \\ 
\hline 
Uniform Masking~\cite{li2022uniform} & proposed uniform masking & all, 25\%mask & pixel & ImageNet-1K (200) & MSE & PVT-S & ViT & 82.04\% & 22.05.20 & arxiv \\
& & & & & & Swin-T & & 82.0\% & & \\ 
\hline 
MixMIM~\cite{liu2022mixmim} & No mask & A mixture of two images & Two image pixel & ImageNet-1K (300) & MSE & ViT-B & 8 blocks & 83.2\% & 22.05.26 & arxiv \\
& & & & ImageNet-1K (300) & & ViT-L & & 85.0\% & & \\ 
& & & & ImageNet-1K (600) & & Swin-B & & 84.4\% & & \\ 
& & & & ImageNet-1K (600) & & Swin-L & & 85.7\% & & \\ 
& & & & ImageNet-1K (300) & & PVT-L & & 83.2\% & & \\ 
\hline 
HiViT~\cite{zhang2022hivit} & random & visible & pixel & ImageNet-1K (300) & MSE & (proposed) HiViT-B & 6 blocks & 83.8\% & 22.05.30 & arxiv \\
& & & & ImageNet-1K (800) & & & & 84.2\% & & \\ 

\hline 
\textbf{MAE augmentor/Others} \\ 
\hline 
CIM~\cite{fang2022corrupted} & reconed=corrupted & all corrupted image & all pixel & ImageNet-1K & L1+L2+discriminative & ViT-S & none & 81.6\% & 22.02.07 & arxiv \\
& & & & & Generator: small BEiT & ViT-B & & 83.3\% & & \\ 
& & & & & & ResNet-50 & & 80.4\% & & \\ 
\hline 
MRA~\cite{xu2022masked} & n & n & n & n & n & MAE & MAE & Incomparable & 22.06.10 & arxiv \\

\hline 
\textbf{Medical images} \\
\hline 
Self-MAE~\cite{zhou2022self} & random & visible & pixel  & ChestX-ray14 (224)  & MAE &  ViT-B &  ViT&  lung disease classification 81.5\% &2022.3.10 & arxiv  \\
\hline 
MIM-Medical~\cite{chen2022masked}   &random   & all  &  raw voxel &  BTCV, TCIA-COVID19 (9)  & l1,l2  & ViT3D-L & linear layer     &  multi-organ segmentation 76.03\%   & 2022.04.25 & arxiv\\
\hline 
DAMA~\cite{ly2022student}   & adaptive  &visible  &  pixel,feature &  ImageNet-1K(224)  & MSE\textcolor{red}{?}  & ViT-B & \textcolor{red}{?}     &  500 epoch, 83.17 & 2022.5.10 & arxiv \\
\hline 
GCMAE ~\cite{quan2022global}   &random  &visible  &  pixel &  Camelyon16(224)  & MSE  & ViT & 8 blocks transformer     &  83.29\%  & 2022.5.18 & arxiv\\
\hline 
SD-MAE ~\cite{luo2022self}   &random  & visible  &  pixel &  PatchCamelyon,NCT-CRC-HE,ImageNet-100 (224)& MSE  & ViT-S & 4  blocks  192d, transformer   &  ImageNet-100,84.63\% & 2022.03.21 & arxiv\\
\hline 
MAE-MIL ~\cite{an2022masked}   &random   &visible  &  pixel &  Camelyon16(1024)  & MSE  & \textcolor{red}{?} & \textcolor{red}{?}     &  61\% & / & MIDL 2022\\
\hline 
\textbf{Multi-modal}  \\
\hline 
MultiMAE~\cite{bachmann2022multimae}   &random  &visible  &  pixel &   ImageNet-1K(224)  & loss  & ViT-B &      &  83.3\% &2022.04.04 & arxiv\\
\hline 
\end{tabular}}
\end{table*}

\begin{table*}[htb]\centering
\vspace{-10pt}
\caption{Summary of works with masked autoencoder on videos.}
\label{tab:summary_of_models_video}
\resizebox{1.0\textwidth}{!}{ 
\begin{tabular}{cccccccccccccccc}
\hline 
model  &masking strategy &input type & prediction target  & pretrain dataset   & image size & Test Set  & encoder & decoder     & Finetune Accuracy & arxiv time & publish status \\
\hline 
BEVT ~\cite{wang2022bevt}  & block-wise/tube & all patches & token & ImageNet-1K, HowTo100M & 224 & SSv2 & Video Swin-Base &  CNN-1, Linear-1 &  70.6\% & 2021.12.02 & CVPR 2022\\
&&&&&& Diving48 & &&86.7\% \\
&&&&&& K400 & &&80.6\% \\
\hline 
MaskFeat ~\cite{wei2022masked} &   cube masking   & all patches &  features (HOG) &  K400&224 &K400 & MViT-S& None & 82.2\% & 2021.12.16 &CVPR 2022 \\
&&&&&&&MViT-L & & 84.3\% \\
\hline 
VideoMAE ~\cite{tong2022videomae}   & tube masking & visible & pixel &SSv2 &  224 & SSv2 &  ViT-B & 4 block, 384d & 69.3\%  \\
&&&&K400&&K400 &&& 79.45\%  &2022.03.23 & arxiv\\
\hline 
MAE-video ~\cite{feichtenhofer2022masked}  & random & visible & pixel &K400 & 224 &K400 & ViT-L & 4 block, 512d& 84.8\%  & 2022.05.18  &arxiv\\
&&&&&&SSv2&& &  72.1\%\\
\hline
OmniMAE~\cite{girdhar2022omnimae} & random & all patches & pixel  &  ImageNet-1K, SSv2  & 224& ImageNet-1K & ViT-B & 4 block, 384d & 82.8\% & 2022.06.16& arxiv\\
&&&&&&SSv2&ViT-B&&69.0\%  \\
&&&&&&ImageNet-1K&ViT-L&&84.7\%  \\
&&&&&&SSv2&ViT-L&&73.4\%  \\
\hline 
\end{tabular}}
\end{table*}

\begin{table*}[htb]\centering
\vspace{-10pt}
\caption{Summary of works with masked autoencoder on point cloud and graph.}
\label{tab:summary_of_mutimodal}
\resizebox{1.0\textwidth}{!}{ 
\begin{tabular}{cccccccccccccccc}
\hline
model  & masking strategy &input type & prediction target & encoder & decoder & pretrain dataset & Finetune Accuracy & arxiv time & publish status \\
\hline
\textbf{Point Cloud} \\
\hline
Point-BERT~\cite{wang2022bevt} & random & all tokens & Tokens extracted from tokenizer (dVAE) & Standard Transformer encoder~\cite{vaswani2017attention} & PointNet (MLP layers)~\cite{qi2017pointnet} & ModelNet40 & 93.8\% & 21.11.29 & CVPR 2022\\
& & & & & & ScanObjectNN & OBJ-BG:87.43, OBJ-ONLY:88.12, PB-T50-RS:83.07 & & \\
& & & & & & ShapeNetPart & 84.11 mIoUc, 85.6 mIoUi & & \\

\hline 
Point-MAE~\cite{pang2022masked} & random & visible tokens & Raw point patches &  Standard transformer encoder & Standard transformer decoder & ModelNet40 & 93.8\% & 220313 & arXiv\\
& & & & & & ScanObjectNN & OBJ-BG:90.02, OBJ-ONLY:88.29, PB-T50-RS:85.18 & & \\ 
& & & & & & ShapeNetPart & 86.1 mIoUi & & \\

\hline 
MaskPoint~\cite{liu2022masked} & random & Encoder: visible tokens & Real or Fake & Standard transformer encoder & Standard transformer decoder & ModelNet40 & 93.8\% & 220321 & arXiv\\ 
& & Decoder: real/fake points & & & & ScanObjectNN & OBJ-BG:88.1, OBJ-ONLY:89.3, PB-T50-RS:84.3 & & \\ 
& & & & & & ShapeNetPart & 84.4 mIoUc, 86.0 mIoUi  & & \\ 

\hline 
Point-M2AE~\cite{zhang2022point} & multi-scale, random & visible tokens & Raw point patches &  Proposed hierarchical transformer & Proposed hierarchical transformer & ModelNet40 & 94.0\% & 220328 & arXiv\\ 
& & & & & & ScanObjectNN & OBJ-BG:91.22, OBJ-ONLY:88.81, PB-T50-RS:86.43 & & \\ 
& & & & & & ShapeNetPart & 84.86 mIoUc, 86.51 mIoUi  & & \\ 
\hline
\textbf{Graph}& & & & & & \textbf{Node classification} & & & \\ 
\hline
MGAE~\cite{tan2022mgae} & random edge & visible & masked edges &  SAGE~\cite{hamilton2017graphsage},GCN~\cite{kipf2017gcn} & dot product, MLP & Cora, CiteSeer, PubMed & 86.15\%, 74.60\%, 86.91\% & 220107 & arXiv\\
& & & & & & (classify 70\% masked edge) & & & \\
\hline
GMAE~\cite{chen2022graph} & random node & visible & masked nodes & Graph Transformers & Graph Transformers & Cora, CiteSeer, PubMed & 81.14\%, 69.25\%, 81.40\% & 220217 & arXiv\\
\hline
GraphMAE~\cite{hou2022graphmae} & random node & all graph & target & GNN(GCN~\cite{kipf2017gcn}, GAT~\cite{velivckovic2017gan}, GIN~\cite{xu2018gin}) & GNN(GAT, GIN) & Cora, CiteSeer, PubMed & 84.2\%, 73.4\%, 81.1\% & 220522 & KDD'22\\
\hline
MaskGAE~\cite{li2022maskgae} & random edge & visible & masked edges, degree regression & GCN & MLP & Cora, CiteSeer, PubMed & 84.05\%, 73.49\%, 83.06\% & 220520 & arXiv\\
\hline
\end{tabular}}
\end{table*}

\end{document}